\documentclass[10pt,twocolumn,letterpaper]{article}

\usepackage{cvpr}
\usepackage{times}
\usepackage{epsfig}
\usepackage{graphicx}
\usepackage{amsmath}
\usepackage{amssymb}
\usepackage{booktabs}
\usepackage{subcaption}
\usepackage{multirow}
\usepackage{balance}

\usepackage[pagebackref=true,breaklinks=true,letterpaper=true,colorlinks,bookmarks=false]{hyperref}

\cvprfinalcopy 

\def\cvprPaperID{****} 

\ifcvprfinal\fi
\begin{document}

\title{Convolutional Mesh Regression for Single-Image Human Shape Reconstruction}

\author{Nikos Kolotouros, Georgios Pavlakos, Kostas Daniilidis \\[0ex]
University of Pennsylvania\\ 
}

\maketitle

\begin{abstract}
This paper addresses the problem of 3D human pose and shape estimation from a single image. Previous approaches consider a parametric model of the human body, SMPL, and attempt to regress the model parameters that give rise to a mesh consistent with image evidence. This parameter regression has been a very challenging task, with model-based approaches underperforming compared to nonparametric solutions in terms of pose estimation. In our work, we propose to relax this heavy reliance on the model's parameter space. We still retain the topology of the SMPL template mesh, but instead of predicting model parameters, we directly regress the 3D location of the mesh vertices. This is a heavy task for a typical network, but our key insight is that the regression becomes significantly easier using a Graph-CNN. This architecture allows us to explicitly encode the template mesh structure within the network and leverage the spatial locality the mesh has to offer. Image-based features are attached to the mesh vertices and the Graph-CNN is responsible to process them on the mesh structure, while the regression target for each vertex is its 3D location. Having recovered the complete 3D geometry of the mesh, if we still require a specific model parametrization, this can be reliably regressed from the vertices locations. We demonstrate the flexibility and the effectiveness of our proposed graph-based mesh regression by attaching different types of features on the mesh vertices. In all cases, we outperform the comparable baselines relying on model parameter regression, while we also achieve state-of-the-art results among model-based pose estimation approaches.
\footnote{Project Page: \url{seas.upenn.edu/~nkolot/projects/cmr}}
\end{abstract}

\section{Introduction}
\begin{figure}[!t]
\centering
\begin{subfigure}[]{.3\columnwidth}
	\includegraphics[width=\columnwidth]{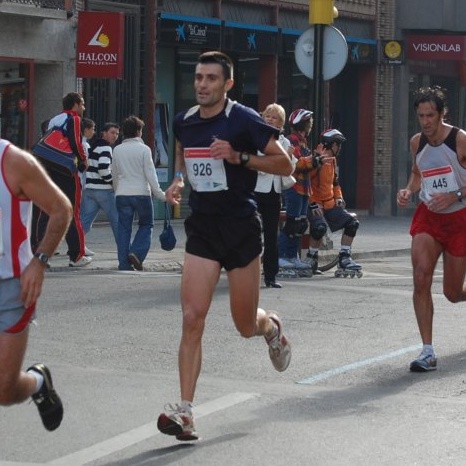}
\end{subfigure}
\begin{subfigure}[]{.3\columnwidth}
	\includegraphics[width=\columnwidth]{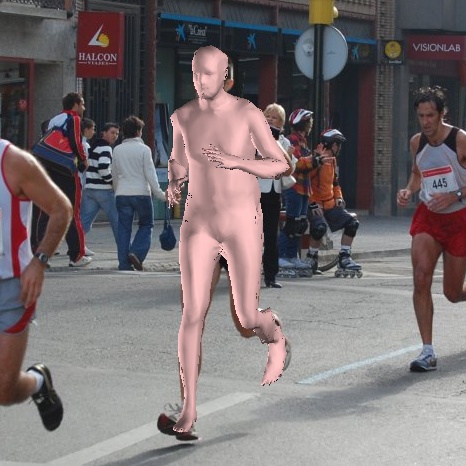}
\end{subfigure}
\begin{subfigure}[]{.3\columnwidth}
	\includegraphics[width=\columnwidth]{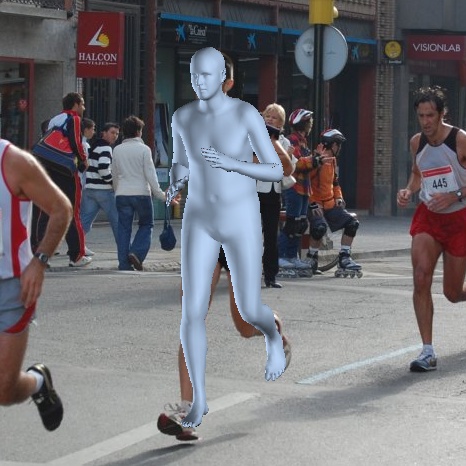}
\end{subfigure}\\
\begin{subfigure}[]{.3\columnwidth}
	\includegraphics[width=\columnwidth]{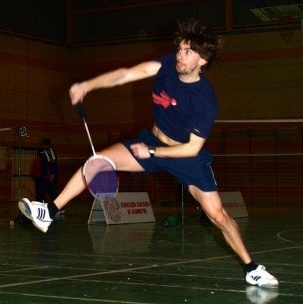}
\end{subfigure}
\begin{subfigure}[]{.3\columnwidth}
	\includegraphics[width=\columnwidth]{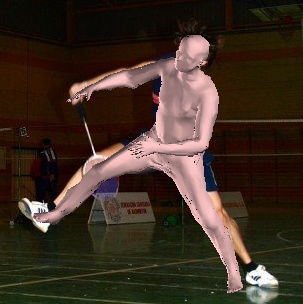}
\end{subfigure}
\begin{subfigure}[]{.3\columnwidth}
	\includegraphics[width=\columnwidth]{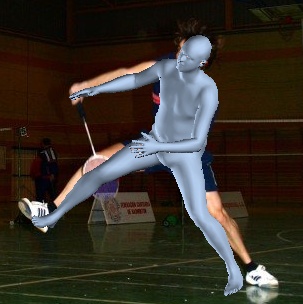}
\end{subfigure}
\caption{Summary of our approach. Given an input image we directly regress a 3D shape with graph convolutions. Optionally, from the 3D shape output we can regress the parametric representation of a body model.}
    \label{fig:figure_teaser}
\end{figure}

\begin{figure*}[!t]
    \centering
    \includegraphics[scale=.65]{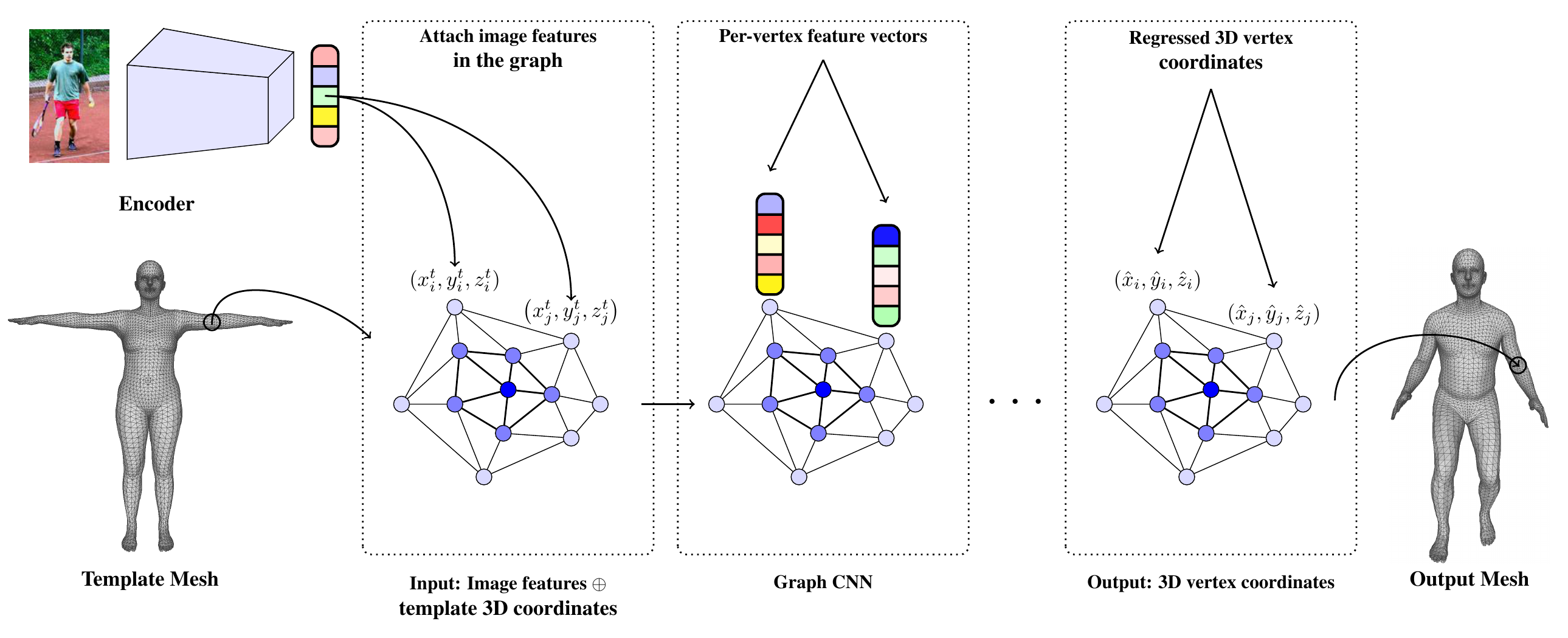}
    \caption{Overview of proposed framework. Given an input image, an image-based CNN encodes it in a low dimensional feature vector. This feature vector is embedded in the graph defined by the template human mesh by attaching it to the 3D coordinates $(x^t_i,y^t_i,z^t_i)$ of every vertex $i$. We then process it through a series of Graph Convolutional layers and regress the 3D vertex coordinates $(\hat{x}_i,\hat{y}_i,\hat{z}_i)$ of the deformed mesh.}
    \label{fig:main_figure}
\end{figure*}

Analyzing humans from images goes beyond estimating the 2D pose for one person~\cite{newell2016stacked, wei2016convolutional} or multiple people~\cite{cao2016realtime, pishchulin2016deepcut}, or even estimating a simplistic 3D skeleton~\cite{martinez2017simple,mehta2017vnect}. Our understanding relies heavily on being able to properly reconstruct the complete 3D pose and shape of people from monocular images. And while this problem is well addressed in settings with multiple cameras~\cite{huang2017towards,joo2018total}, the excessive ambiguity, the limited training data, and the wide range of imaging conditions make this task particularly challenging in the monocular case.

Traditionally, optimization-based approaches~\cite{bogo2016keep,lassner2017unite,zanfir2018monocular} have offered the most reliable solution for monocular pose and shape recovery. However, the slow running time, the reliance on a good initialization and the typical failures due to bad local minima have recently shifted the focus to learning-based approaches~\cite{kanazawa2018end,lassner2017unite, omran2018neural,pavlakos2018learning,tan2017indirect,tung2017self}, that regress pose and shape directly from images. The majority of these works investigate what is the most reliable modality to regress pose and shape from. Surface landmarks~\cite{lassner2017unite}, pose keypoints and silhouettes~\cite{pavlakos2018learning}, semantic part segmentation~\cite{omran2018neural}, or raw pixels~\cite{kanazawa2018end} have all been considered as the network input. And while the input representation topic has received much debate, all the above approaches nicely conform to the SMPL model~\cite{loper2015smpl} and use its parametric representation as the regression target of choice. However, taking the decision to commit to a particular parametric space can be quite constraining itself. For example, SMPL is not modeling hand pose or facial expressions~\cite{joo2018total,romero2017embodied}. What is even more alarming is that the model parameter space might not be appropriate as a regression target. In the case of SMPL, the pose space is expressed in the form of 3D rotations, a pretty challenging prediction target~\cite{mahendran2018mixed,mousavian20173d}. Depending on the selected 3D rotation representation (e.g., axis angle, rotation matrices, quaternions), we might face problems of periodicity, non-minimal representation, or discontinuities, which complicate the prediction task. And in fact, all the above model-based approaches underperfom in pose estimation metrics compared to approaches regressing a less informative, yet more accurate, 3D skeleton through 3D joint regression~\cite{dabral2018learning,martinez2017simple,pavlakos2018ordinal,sun2017integral}.

In this work, we propose to take a more hybrid route towards pose and shape regression. Even though we preserve the template mesh introduced by SMPL, we do not directly regress the SMPL model parameters. Instead, our regression target is the 3D mesh vertices. Considering the excessive number of vertices of the mesh, if addressed naively, this would be a particular heavy burden for the network. Our key insight though, is that this task can be effectively and efficiently addressed by the introduction of a Graph-CNN. This architecture enables the explicit encoding of the mesh structure in the network, and leverages the spatial locality of the graph. Given a single image (Figure~\ref{fig:main_figure}), any typical CNN can be used for feature extraction. The extracted features are attached on the vertex coordinates of the template mesh, and the processing continues on the graph structure defined for the Graph-CNN. In the end, each vertex has as target its 3D location in the deformed mesh. This allows us to recover the complete 3D geometry of the human body without explicitly committing to a pre-specified parametric space, leaving the mesh topology as the only hand-designed choice. Conveniently, after estimating the 3D position for each vertex, if we need our prediction to conform to a specific model, we can regress its parameters quite reliably from the mesh geometry  (Figure~\ref{fig:figure_teaser}). This enables a more hybrid usage for our approach, making it directly comparable to model-based approaches. Furthermore, our graph-based processing is largely agnostic to the input type, allowing us to attach features extracted from RGB pixels~\cite{kanazawa2018end}, semantic part segmentation~\cite{omran2018neural}, or even from dense correspondences~\cite{guler2018densepose}. In all these cases we demonstrate that our approach outperforms the baselines that regress model parameters directly from the same type of features, while overall we achieve state-of-the-art pose estimation results among model-based baselines.

Our contributions can be summarized as follows:
\begin{itemize}
\item We reformulate the problem of human pose and shape estimation in the form of regressing the 3D locations of the mesh vertices, to avoid the difficulties of direct model parameter regression. 
\item We propose a Graph CNN for this task which encodes the mesh structure and enables the convolutional mesh regression of the 3D vertex locations.
\item We demonstrate the flexibility of our framework by considering different input representations, always outperforming the baselines regressing the model parameters directly.
\item We achieve state-of-the-art results among model-based pose estimation approaches.
\end{itemize}

\section{Related work}
There is rich recent literature on 3D pose estimation in the form of a simplistic body skeleton, e.g.,~\cite{dabral2018learning,li20143d,luvizon20182d,martinez2017simple,mehta2017vnect,
pavlakos2018ordinal,pavlakos2017coarse,rogez2016mocap,rogez2017lcr,sun2017integral,
tekin2016structured,tekin2017learning,tome2017lifting,
zhou2017towards,zhou2019monocap}. 
However, in this Section, we focus on the more relevant works recovering 
the full shape and pose of the human body.

\noindent
\textbf{Optimization-based shape recovery}:
Going beyond a simplistic skeleton, and recovering the full pose and shape, initially, the most successful approaches followed optimization-based solutions. The work of Guan~\etal~\cite{guan2009estimating} relied on annotated 2D landmarks and optimized for the parameters of the SCAPE parametric model that generated a mesh optimally matching this evidence. This procedure was made automatic with the SMPLify approach of Bogo~\etal~\cite{bogo2016keep}, where the 2D keypoints where localized through the help of a CNN~\cite{pishchulin2016deepcut}. Lassner~\etal~\cite{lassner2017unite} included auxiliary landmarks on the surface of the human body, and additionally considered the estimated silhouette during the fitting process. Zanfir~\etal~\cite{zanfir2018monocular} similarly optimized for consistency of the reprojected mesh with semantic parts of the human body, while extending the approach to work for multiple people as well. Despite the reliable results obtained, the main concern for approaches of this type is that they pose a complicated non-convex optimization problem. This means that the final solution is very sensitive to the initialization, the optimization can get stuck in local minima, and simultaneously the whole procedure can take several minutes to complete. These drawbacks have motivated the increased interest in learning-based approaches, like ours, where the pose and shape are regressed directly from images.

\noindent
\textbf{Direct parametric regression}:
When it comes to pose and shape regression, the vast majority of works adopt the SMPL parametric model and consider regression of pose and shape parameters. Lassner~\etal~\cite{lassner2017unite} detect 91 landmarks on the body surface and use a random forest to regress the SMPL model parameters for pose and shape. Pavlakos~\etal~\cite{pavlakos2018learning} rely on a smaller number of keypoints and body silhouettes to regress the SMPL parameters. Omran~\etal~\cite{omran2018neural} follow a similar strategy but use a part segmentation map as the intermediate representation. On the other hand, Kanazawa~\etal~\cite{kanazawa2018end} attempt to regress the SMPL parameters directly from images, using a weakly supervised approach relying on 2D keypoint reprojection and a pose prior learnt in an adversarial manner. Tung~\etal~\cite{tung2017self} present a self-supervised approach for the same problem, while Tan~\etal~\cite{tan2017indirect} rely on weaker supervision in the form of body silhouettes. The common theme of all these works is that they have focused on using the SMPL parameter space as a regression target. However, the 3D rotations involved as the pose parameters have created issues in the regression (e.g., discontinuities or periodicity) and typically underperform in terms of pose estimation compared to skeleton-only baselines. In this work, we propose to take an orthogonal approach to them, by regressing the 3D location of the mesh vertices by means of a Graph-CNN. Our approach is transparent to the type of the input representation we use, since the flexibility of the Graph network allows us to consider different types of input representations employed in prior work, like semantic part-based features~\cite{omran2018neural}, features extracted directly from raw pixels~\cite{kanazawa2018end}, or even dense correspondences~\cite{guler2018densepose}.

\noindent
\textbf{Nonparametric shape estimation}:
Recently, nonparametric approaches have also been proposed for pose and shape estimation. Varol~\etal~\cite{varol2018bodynet} use a volumetric reconstruction approach with a voxel output. Different tasks are simultaneously considered for intermediate supervision. Jackson~\etal~\cite{jackson20183d} also propose a form of volumetric reconstruction by extending their recent face reconstruction network~\cite{jackson2017large} to work for full body images. The main drawback of these approaches adopting a completely nonparametric route, is that even if they recover an accurate voxelized sculpture of the human body, there is none or very little semantic information captured. In fact, to recover the body pose, we need to explicitly perform an expensive body model fitting step using the recovered voxel map, as done in~\cite{varol2018bodynet}. In contrast to them, we retain the SMPL mesh topology, which allows us to get dense semantic correspondences of our 3D prediction with the image, and in the end we can also easily regress the model's parameters given the vertices 3D location.

\noindent
\textbf{Graph CNNs}:
Wang~\etal~\cite{wang2018eccv} use a Graph CNN to reconstruct meshes of objects from images by deforming an initial ellipsoid. However, mesh reconstruction of arbitrary objects is still an open problem, because shapes of objects even in the same class, e.g., chairs, do not have the same genus. Contrary to generic objects, arbitrary human shapes can be reconstructed as continuous deformations of a template model. In fact, recently there has been a lot of research in applying Graph Convolutions for human shape applications. Verma~\etal~\cite{verma2018feastnet} propose a new data-driven Graph Convolution operator with applications on shape analysis. Litany~\etal~\cite{litany2017deformable} use a Graph VAE to learn a latent space of human shapes, that is useful for shape completion. Ranjan~\etal~\cite{ranjan2018generating} use a mesh autoencoder network to recover a latent representation of 3D human faces from a series of meshes. The main difference of our approach is that we do not aim to learn a generative shape model from 3D shapes, but instead perform single-image shape reconstruction; the input to our network is an image, not a 3D shape. The use of a Graph CNN alone is not new, but we consider as a contribution the insight that Graph CNNs provide a very natural structure to enable our hybrid approach. They assist us in avoiding the SMPL parameter space, which has been reported to have issues with regression~\cite{martinez2017simple, pavlakos2018learning}, while simultaneously allowing the explicit encoding of the graph structure in the network, so that we can leverage spatial locality and preserve the semantic correspondences.

\section{Technical approach}
In this Section we present our proposed approach for predicting 3D human shape from a single image.  First, in Subsection~\ref{sec:image_cnn} we briefly describe the image-based architecture that we use as a generic feature extractor. In Subsection~\ref{sec:graph_cnn} we focus on the core of our approach, the Graph CNN architecture that is responsible to regress the 3D vertex coordinates of the mesh that deforms to reconstruct the human body. Then, Subsection~\ref{sec:s2s} describes a way to combine our non-parametric regression with the prediction of SMPL model parameters. Finally, Subsection~\ref{sec:details} focuses on important implementation details.

\subsection{Image-based CNN}\label{sec:image_cnn}
The first part of our pipeline consists of a typical image-based CNN following the ResNet-50 architecture~\cite{he2016resnet}. From the original design we ignore the final fully connected layer, keeping only the 2048-D feature vector after the average pooling layer. This CNN is used as a generic feature extractor from the input representation. To demonstrate the flexibility of our approach, we experiment with a variety of inputs, i.e., RGB images, part segmentation and DensePose input~\cite{guler2018densepose}. For RGB images we simply use raw pixels as input, while for the other representations, we assume that another network~\cite{guler2018densepose}, provides us with the predicted part segmentation or DensePose. Although we present experiments with a variety of inputs, our goal is not to investigate the effect of the input representation, but rather we focus our attention on the graph-based processing that follows.

\subsection{Graph CNN}\label{sec:graph_cnn}
At the heart of our approach, we propose to employ a Graph CNN to regress the 3D coordinates of the mesh vertices. For our network architecture we draw inspiration from the work of Litany~\etal~\cite{litany2017deformable}. We start from a template human mesh with $N$ vertices as depicted in \figurename~\ref{fig:main_figure}. Given the 2048-D feature vector extracted by the generic image-based network, we attach these features to the 3D coordinates of each vertex in the template mesh. From a high-level perspective, the Graph CNN uses as input the 3D coordinates of each vertex along with the input features and has the goal of estimating the 3D coordinates for each vertex in the output, deformed mesh. This processing is performed by a series of Graph Convolution layers.

For the graph convolutions we use the formulation from Kipf~\etal~\cite{kipf2017semi} which is defined as:
\begin{equation}
    Y = \tilde{A}XW
\end{equation}
where $X \in \mathbb{R}^{N\times k}$ is the input feature vector, $W \in \mathbb{R}^{k\times \ell}$ the weight matrix and $\tilde{A} \in \mathbb{R}^{N\times N}$ is the row-normalized adjacency matrix of the graph. Essentially, this is equivalent to performing per-vertex fully connected operations followed by a neighborhood averaging operation. The neighborhood averaging is essential for producing a high quality shape because it enforces neighboring vertices to have similar features, and thus the output shape is smooth. With this design choice we observed that there is no need of a smoothness loss on the shape, as for example in~\cite{kato2018renderer}. We also experimented with the more powerful graph convolutions proposed in~\cite{verma2018feastnet} but we did not observe quantitative improvement in the results, so we decided to keep our original and simpler design choice.

For the graph convolution layers, we make use of residual connections as they help in speeding up significantly the training and also lead in higher quality output shapes. Our basic building block is similar to the Bottleneck residual block~\cite{he2016resnet} where $1\times 1$ convolutions are replaced by per-vertex fully connected layers and Batch Normalization~\cite{ioffe2015icml} is replaced by Group Normalization~\cite{wu2018eccv}. We noticed that Batch Normalization leads to unstable training and poor test performance, whereas with no normalization the training is very slow and the network can get stuck at local minima and collapse early during training. 

Besides the 3D coordinates for each vertex, our Graph CNN also regresses the camera parameters for a weak-perspective camera model. Following Kanazawa~\etal~\cite{kanazawa2018end}, we predict a scaling factor $s$ and a 2D translation vector $\mathbf{t}$. Since the prediction of the network is already on the camera frame, we do not need to regress an additional global camera rotation. The camera parameters are regressed from the graph embedding and not from the image features directly. This way we get a much more reliable estimate that is consistent with the output shape. 

Regarding training, let $\hat{Y} \in \mathbb{R}^{N \times 3}$ be the predicted 3D shape, $Y$ the ground truth shape and $X$ the ground truth 2D keypoint locations of the joints. From our 3D shape we can also regress the location for the predicted 3D joints $\hat{J}_{3D}$ employing the same regressor that the SMPL model is using to recover joints from vertices. Given these 3D joints, we can simply project them on the image plane, $\hat{X} = s\mathbf{\Pi}(\hat{J}_{3D}) + \mathbf{t}$. Now, we train the network using two forms of supervision. First, we apply a per-vertex $L_1$ loss between the predicted and ground truth shape, i.e.,
\begin{equation}
    \mathcal{L}_{shape} = \sum_{i=1}^N||\hat{Y}_i-Y_i||_{1}.
\end{equation}
Empirically we found that using $L_1$ loss leads to more stable training and better performance than $L_2$ loss. Additionally, to enforce image-model alignment, we also apply an $L_1$ loss between the projected joint locations and the ground truth keypoints, i.e.,
\begin{equation}
    \mathcal{L}_{J} = \sum_{i=1}^M||\hat{X}_{i} - X_{i}||_{1}.
\end{equation}
Finally, our complete training objective is:
\begin{equation}
    \mathcal{L} = \mathcal{L}_{shape} + \mathcal{L}_{J}.
\end{equation}

This form of supervised training requires us to have access to images with full 3D ground truth shape. However, based on our empirical observation, it is not necessary for all the training examples to come with ground truth shape. In fact, following the observation of Omran~\etal~\cite{omran2018neural}, we can leverage additional images that provide only 2D keypoint ground truth. In these cases, we simply ignore the first term of the previous equation and train only with the keypoint loss. We have included evaluation under this setting of weaker supervision in the Sup.~Mat.

\subsection{SMPL from regressed shape}\label{sec:s2s}
\begin{figure}[!t]
    \centering
    \includegraphics[scale=.7]{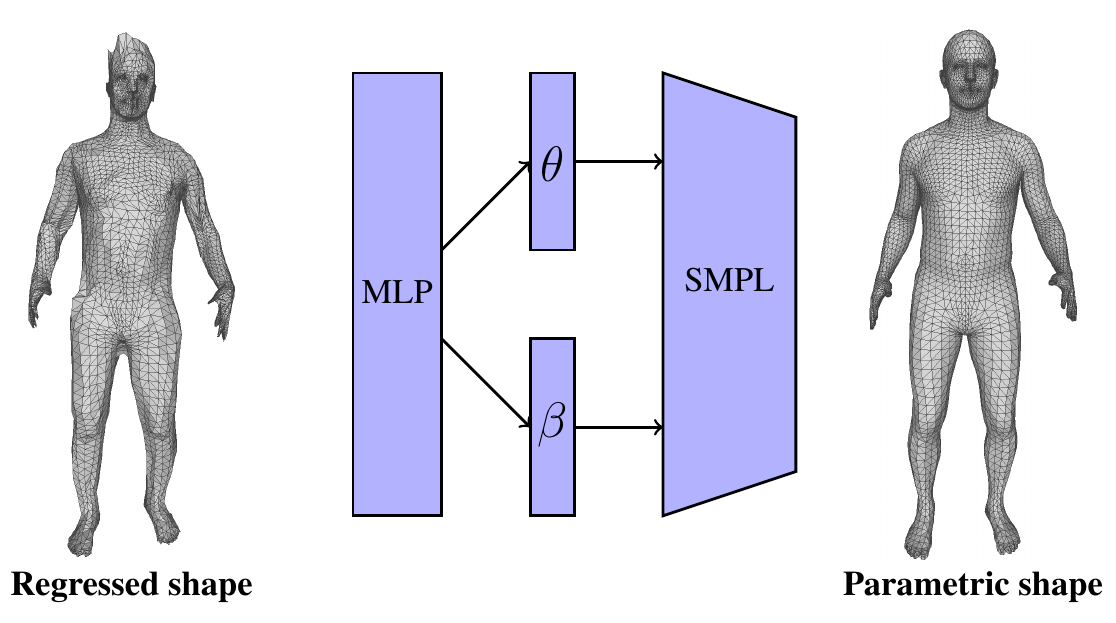}
    \caption{\textbf{Predicting SMPL parameters from regressed shape}. Given a regressed 3D shape from the network of \figurename~\ref{fig:main_figure}, we can use a Multi-Layer Perceptron (MLP) to regress the SMPL parameters and produce a shape that is consistent with the original non-parametric shape}
    \label{fig:shape_to_smpl}
	\vspace*{-1.0em}
\end{figure}
Although we demonstrate that non-parametric regression is an easier task for the network, there are still many applications where a parametric representation of the human body can be very useful (e.g., motion prediction). In this Subsection, we present a straightforward way to combine our non-parametric prediction with a particular parametric model, i.e., SMPL. To achieve this goal, we train another network that regresses pose ($\theta$) and shape ($\beta$) parameters of the SMPL parametric model given the regressed 3D shape as input. The architecture of this network can be very simple, i.e., a Multi-Layer Perceptron (MLP)~\cite{rumelhart86} for our implementation. This network is presented in \figurename~\ref{fig:shape_to_smpl} and the loss function for training is:
\begin{equation}
    \mathcal{L} = \mathcal{L}_{shape} + \mathcal{L}_{J} + \mathcal{L}_{\theta} + \lambda \mathcal{L}_{\beta}.
\end{equation}
Here, $\mathcal{L}_{shape}$ and $\mathcal{L}_{J}$ are the losses on the 3D shape and 2D joint reprojection as before, while $\mathcal{L}_{\theta}$ and $\mathcal{L}_{\beta}$ are $L_2$ losses on the SMPL pose and shape parameters respectively.

As observed by previous works, e.g.,~\cite{pavlakos2018learning, martinez2017simple}, it is challenging to regress the pose parameters $\theta$, which represent 3D rotations in the axis-angle representation. To avoid this, we followed the strategy employed by Omran~\etal~\cite{omran2018neural}. More specifically, we convert the parameters from axis-angle representation to a rotation matrix representation using the Rodrigues formula, and we set the output of our network to regress the elements of the rotation matrices. To ensure that the output is a valid rotation matrix we project it to the manifold of rotation matrices using the differentiable SVD operation. Although this representation does not explicitly improve our quantitative results, we observed faster convergence during training, so we selected it as a more practical option.

\subsection{Implementation details}\label{sec:details}

An important detail regarding our Graph CNN is that we do not operate directly on the original SMPL mesh, but we first subsample it by a factor of 4 and then upsample it again to the original scale using the technique described in \cite{ranjan2018generating}. This is essentially performed by precomputing downsampling and upsampling matrices $D$ and $U$ and left-multiply them with the graph every time we need to do resampling. This downsampling step helps to avoid the high redundancy in the original mesh due to the spatial locality of the vertices, and decrease memory requirements during training.

Regarding the training of the MLP, we employ a 2-step training procedure. First we train the network that regresses the non-parametric shape and then with this network fixed we train the MLP that predicts the SMPL parameters. We also experimented with training them end-to-end but we observed a decrease in the performance of the network for both the parametric and non-parametric shape. 

\section{Empirical evaluation}
In this Section, we present the empirical evaluation of our approach. First, we discuss the datasets we use in our evaluation (Subsection~\ref{sec:datasets}), then we provide training details for our pipeline (Subsection~\ref{sec:training}), and finally, the quantitative and qualitative evaluation (Subsection~\ref{sec:evaluation}) follows.

\subsection{Datasets}\label{sec:datasets}
We employ two datasets that provide 3D ground truth for training, Human3.6M~\cite{ionescu2014} and UP-3D~\cite{lassner2017unite}, while we evaluate our approach on Human3.6M and the LSP dataset~\cite{johnson2010clustered}.

\noindent
\textbf{Human3.6M}:
It is an indoor 3D pose dataset including subjects performing activities like Walking, Eating and Smoking. We use the subjects S1, S5, S6, S7 and S8 for training, and keep the subjects S9 and S11 for testing. We present results for two popular protocols (P1 and P2, as defined in~\cite{kanazawa2018end}) and two error metrics (MPJPE and Reconstruction error, as defined in~\cite{zhou2019monocap}).

\noindent
\textbf{UP-3D}:
It is a dataset created by applying SMPLify~\cite{bogo2016keep} on natural images of humans and selecting the successful fits. We use the training set of this dataset for training.

\noindent
\textbf{LSP}:
It is a 2D pose dataset, including also segmentation annotations provided by Lassner~\etal~\cite{lassner2017unite}. We use the test set of this dataset for evaluation.

\subsection{Training details}\label{sec:training}
For the image-based encoder, we use a ResNet50 model~\cite{he2016resnet} pretrained on ImageNet~\cite{imagenet_cvpr09}. All other network components (Graph CNN and MLP for SMPL parameters) are trained from scratch. For our training, we use the Adam optimizer, and a batch size of 16, with the learning rate set to $3\mathrm{e}$\,--\,$4$. We did not use learning rate decay. Training with data only from Human3.6M lasts for 10 epochs, while mixed training with data from Human3.6M and UP-3D requires training for 25 epochs, because of the greater image diversity. To train the MLP that regresses SMPL parameters from our predicted shape, we use 3D shapes from Human3.6M and UP-3D. Finally, for the models using Part Segmentation or DensePose~\cite{guler2018densepose} predictions as input, we use the pretrained network of~\cite{guler2018densepose} to provide the corresponding predictions.

\subsection{Experimental analysis}\label{sec:evaluation}

\begin{table}\centering
\footnotesize
	\begin{tabular}{l|c|c}
	\toprule
	Method & MPJPE & Reconst. Error\\
	\midrule
	SMPL Parameter Regression~\cite{kanazawa2018end} & - & 77.6\\ 
	Mesh Regression (FC) & 200.8 &105.8\\
	Mesh Regression (Graph)	 & {\bf 102.1}  & 69.0\\
	Mesh Regression (Graph + SMPL)	 & 113.2 & {\bf 61.3}\\
	\bottomrule
	\end{tabular}
	\caption{
Evaluation of 3D pose estimation in Human3.6M (Protocol 2). The numbers are MPJPE and Reconstruction errors in mm. Our graph-based mesh regression (with or without SMPL parameter regression) is compared with a method that regresses SMPL parameters directly, as well as with a naive mesh regression using fully connected (FC) layers instead of a Graph-CNN.
}
	\vspace*{-1.0em}
	\label{table:baseline}
\end{table}

\begin{figure}[!ht]
	\centering
	\begin{subfigure}[]{.3\columnwidth}
		\includegraphics[width=\columnwidth]{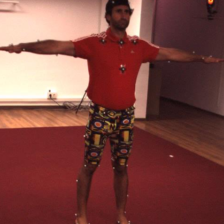}
	\end{subfigure}
	\begin{subfigure}[]{.3\columnwidth}
		\includegraphics[width=\columnwidth]{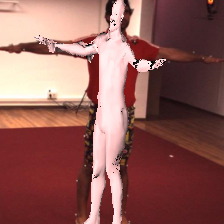}
	\end{subfigure}
	\begin{subfigure}[]{.3\columnwidth}
		\includegraphics[width=\columnwidth]{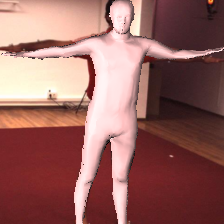}
	\end{subfigure}\\
	\begin{subfigure}[]{.3\columnwidth}
		\includegraphics[width=\columnwidth]{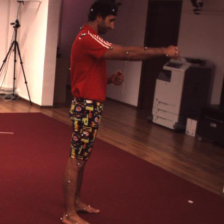}
				\caption*{Image}
	\end{subfigure}
	\begin{subfigure}[]{.3\columnwidth}
		\includegraphics[width=\columnwidth]{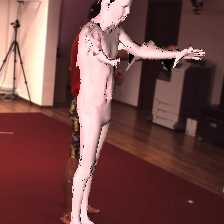}
				\caption*{FC}
	\end{subfigure}
	\begin{subfigure}[]{.3\columnwidth}
		\includegraphics[width=\columnwidth]{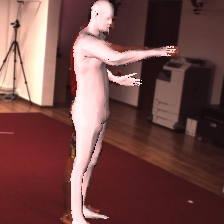}
				\caption*{Graph CNN}
	\end{subfigure}
	\caption{Using a series of fully connected (FC) layers to regress the vertex 3D coordinates severely complicates the regression task and gives non-smooth meshes, since the network cannot leverage directly the topology of the graph.}
	\label{fig:baseline}
	\vspace*{-1.0em}	
\end{figure}

\begin{table}
\footnotesize
\centering
	\begin{tabular}{c|c|cc|cc}
		\toprule
		\multirow{2}{*}{Input} & \multirow{2}{*}{Regression Type} & \multicolumn{2}{|c|}{MPJPE} & \multicolumn{2}{c}{Reconst. Error}
		\\
		& & P1 & P2 & P1 & P2\\
		\midrule
		\multirow{2}{*}{RGB} & Parameter~\cite{kanazawa2018end} & 88.0 & - & 58.1 & 56.8\\
		& Mesh (Graph + SMPL) & {\bf 74.7} & 71.9 & {\bf 51.9} & {\bf 50.1}\\
		\midrule 
		\multirow{2}{*}{Parts} & Parameter~\cite{omran2018neural} & - & - & - & 	59.9 \\
		& Mesh (Graph + SMPL) & 80.4 & 77.4 & 56.1 & {\bf 53.3}\\
		\midrule 
		\multirow{2}{*}{DP\cite{guler2018densepose}} & Parameter~\cite{kanazawa2018end} & 82.7 & 79.5 & 57.8 & 54.9 \\
		& Mesh (Graph + SMPL) & {\bf 78.9} & {\bf 74.2} & {\bf 55.3} & {\bf 51.0}\\ 
		\bottomrule
	\end{tabular}
	\caption{Comparison of direct SMPL parameter regression versus our proposed mesh regression on Human3.6M (Protocol 1 and 2) for different input representations. The numbers are mean 3D joint errors in mm,  with and without Procrustes alignment (Rec. Error and MPJPE respectively). Our results are computed after regressing SMPL parameters from our non-parametric shape. Number are taken from the respective works,  except for the baseline of~\cite{kanazawa2018end} on DensePose images, which is evaluated by us.
	}
	\label{table:diff_inp}
	\vspace*{-1.0em}	
\end{table}

\textbf{Regression target}:
For the initial ablative study, we aim to investigate the importance of our mesh regression for 3D human shape estimation. To this end, we focus on the Human3.6M dataset and we evaluate the regressed shape through 3D pose accuracy. First, we evaluate the direct regression of the 3D vertex coordinates, in comparison to generating the 3D shape implicitly through regression of the SMPL model parameters directly from images. The most relevant baseline in this category is the HMR method of~\cite{kanazawa2018end}. In Table~\ref{table:baseline}, we present the comparison of this approach ({\em SMPL parameter regression}) with our non-parametric shape regression ({\em Mesh Regression - (Graph)}). For a more fair comparison, we also include our results for the MLP that regresses SMPL parameters using our non-parametric mesh as input ({\em Mesh Regression - (Graph + SMPL)}). In both cases, we outperform the strong baseline of~\cite{kanazawa2018end}, which demonstrates the benefit of estimating a more flexible non-parametric regression target, instead of regressing the model parameters in one shot.

Beyond the regression target, one of our contributions is also the insight that the task of regressing 3D vertex coordinates can be greatly simplified when a Graph CNN is used for the prediction. To investigate this design choice, we compare it with a naive alternative that regresses vertex coordinates with a series of fully connected layers on top of our image-based encoder ({\em Mesh Regression - (FC)}). This design clearly underperforms compared to our Graph-based architecture, demonstrating the importance of leveraging the mesh structure through the Graph CNN during the regression. The benefit of graph-based processing is demonstrated also qualitatively in \figurename~\ref{fig:baseline}.

\textbf{Input representation}:
For the next ablative, we demonstrate the effectiveness of our mesh regression for different types of input representations, i.e., RGB images, Part Segmentation as well as DensePose images~\cite{guler2018densepose}. The complete results are presented in~\tablename~{\ref{table:diff_inp}}. The RGB model is trained on Human3.6M + UP-3D whereas the two other models only on Human3.6M. For every input type, we compare with state-of-the-art methods~\cite{kanazawa2018end,omran2018neural} and show that our method outperforms them in all setting and metrics. Interestingly, when training only with Human3.6M data, RGB input performs worse than the other representations (\tablename~{\ref{table:baseline}), because of over-fitting. However, we observed that RGB features capture richer information for in-the-wild images, thus we select it for the majority of our experiments.

\begin{table}
\footnotesize
\centering
	\begin{tabular}{c|c|cc|cc}
		\toprule
		\multirow{2}{*}{Input} & \multirow{2}{*}{Output shape} & \multicolumn{2}{|c|}{MPJPE} & \multicolumn{2}{c}{Reconst. Error}
		\\
		& & P1 & P2 & P1 & P2\\
		\midrule
		\multirow{2}{*}{RGB} & Non parametric & 75.0 & 72.7 & 51.2 & 49.3\\
		& Parametric & 74.7 & 71.9 & 51.9 & 50.1\\
		\midrule 
		\multirow{2}{*}{Parts} & Non parametric & 78.0 & 73.4 & 54.6 & 50.6 \\
		& Parametric & 80.4 & 77.4 & 56.1 & 53.3\\
		\midrule 
		\multirow{2}{*}{DP\cite{guler2018densepose}} & Non parametric
		& 78.0 & 72.3 & 55.3 & 50.3 \\
		& Parametric & 78.9 & 74.2 & 55.3 & 51.0\\ 
		\bottomrule
	\end{tabular}
	\caption{Comparison on Human3.6M (Protocol 1 and 2) of our non-parametric mesh with the SMPL parametric mesh regressed from our shape. Numbers are 3D joint errors in mm. The performance of the two baselines is similar.
}
	\label{table:np_smpl}
	\vspace*{-1.0em}	
\end{table}

\begin{table}\centering
	\begin{tabular}{c|c}
		\toprule
		Method & Reconst. Error\\
		\midrule
		Lassner \etal \cite{lassner2017unite} & 93.9\\
		SMPLify \cite{bogo2016keep} & 	82.3\\
		Pavlakos \etal \cite{pavlakos2018learning} & 75.9\\
		NBF \cite{omran2018neural} & 59.9\\
		HMR \cite{kanazawa2018end} & 56.8\\
		Ours & \textbf{50.1}\\
		\bottomrule
	\end{tabular}
	\caption{Comparison with the state-of-the-art on Human3.6M (Protocol 2). Numbers are Reconstruction errors in mm. Our approach outperforms the previous baselines.}
	\label{table:all}
	\vspace*{-1.0em}	
\end{table}

\begin{table}
	\centering
	\small
	\hspace{-3mm}
	\tabcolsep=2.95mm
	\begin{tabular}{@{}lcccc@{}}
		\toprule
		& \multicolumn{2}{c}{FB Seg.} & \multicolumn{2}{c}{Part Seg.} \\
		\cmidrule{2-5}
		& acc. & f1 & acc. & f1 \\
		\midrule
		SMPLify \emph{oracle}~\cite{bogo2016keep} & 92.17 & 0.88 & 88.82 & 0.67 \\
		SMPLify~\cite{bogo2016keep} & 91.89 & 0.88 & 87.71 & 0.64 \\
		SMPLify on~\cite{pavlakos2018learning} & 92.17 & 0.88 & 88.24 & 0.64 \\
		Bodynet \cite{varol2018bodynet} & 92.75 & 0.84 & - & -\\
		\midrule
		HMR~\cite{kanazawa2018end} & 91.67 & 0.87 & 87.12 & 0.60 \\
		Ours & 91.46 & 0.87 & 88.69 & 0.66 \\
		\bottomrule
	\end{tabular}
	\caption{Segmentation evaluation on the LSP test set. The numbers are accuracies and f1 scores. We include approaches that are purely regression-based (bottom) and approaches that perform some optimization (post)-processing (top). Our approach is competitive with the state-of-the-art.
	}
	\label{tab:lsp}
\vspace*{-1.0em}
\end{table}

\begin{figure}[!t]
	\centering
	\begin{subfigure}[]{.3\columnwidth}
		\includegraphics[width=\columnwidth]{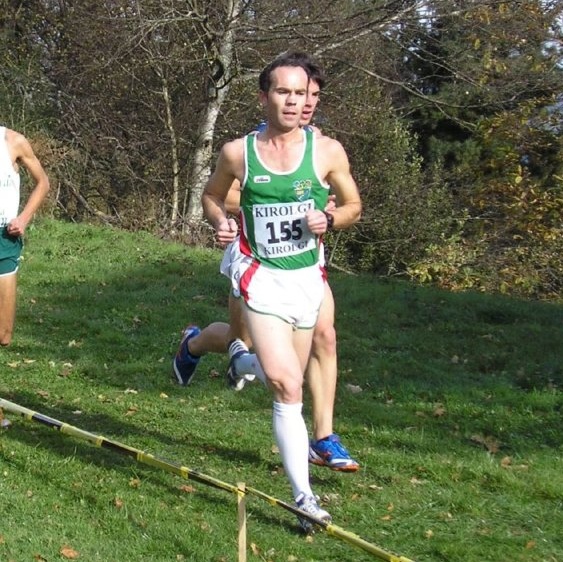}
	\end{subfigure}
	\begin{subfigure}[]{.3\columnwidth}
		\includegraphics[width=\columnwidth]{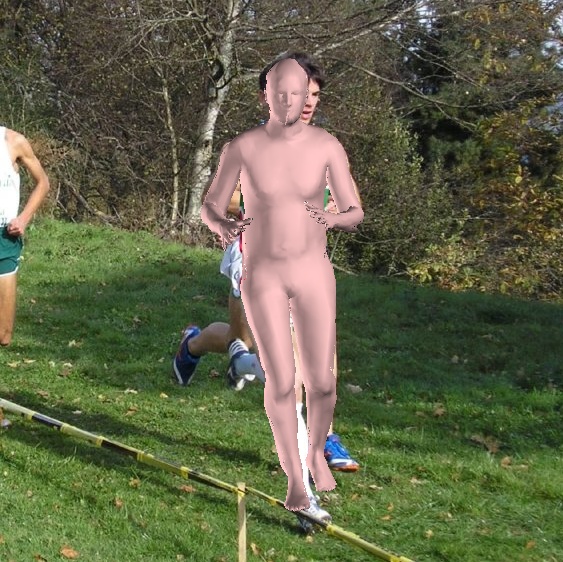}
	\end{subfigure}
	\begin{subfigure}[]{.3\columnwidth}
		\includegraphics[width=\columnwidth]{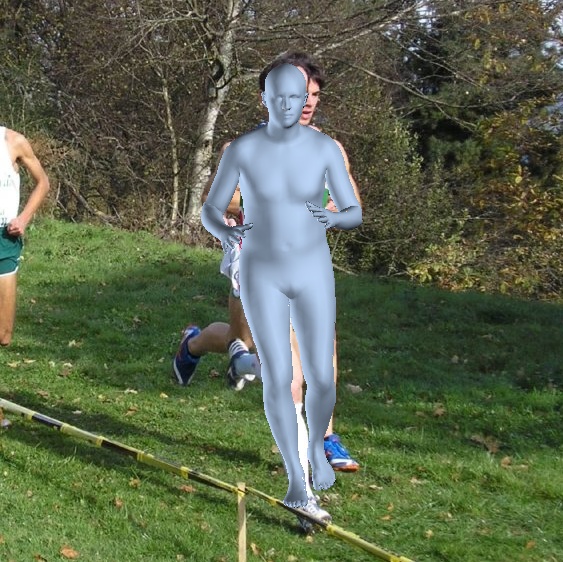}
	\end{subfigure}\\
		\begin{subfigure}[]{.3\columnwidth}
		\includegraphics[width=\columnwidth]{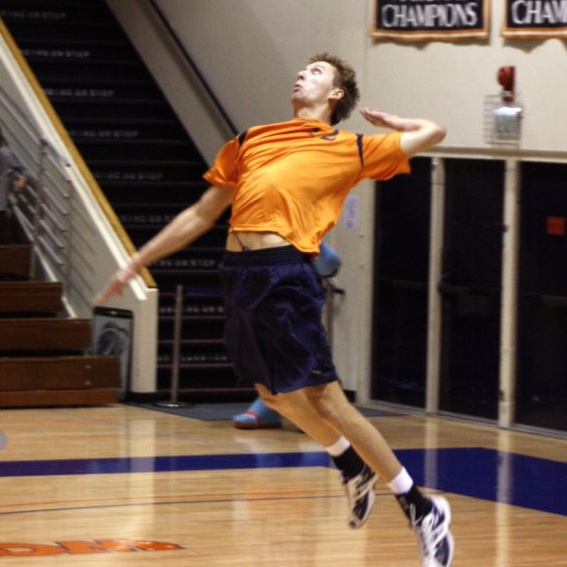}
	\end{subfigure}
	\begin{subfigure}[]{.3\columnwidth}
		\includegraphics[width=\columnwidth]{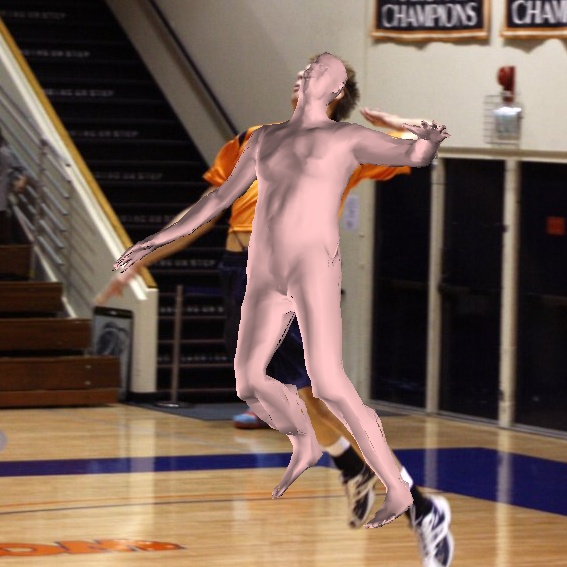}
	\end{subfigure}
	\begin{subfigure}[]{.3\columnwidth}
		\includegraphics[width=\columnwidth]{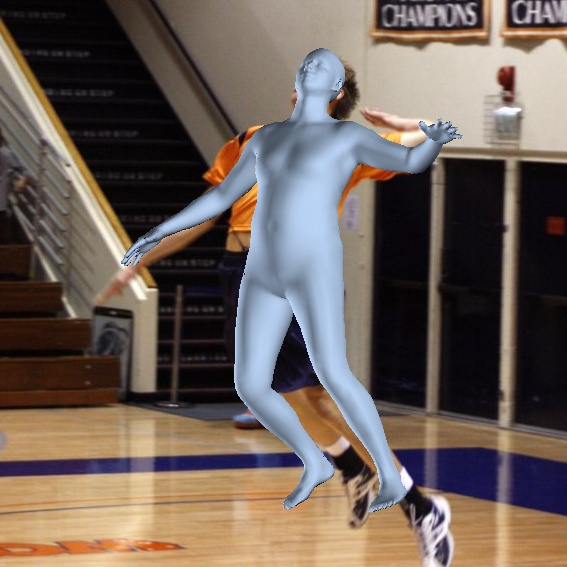}
	\end{subfigure}\\
	\begin{subfigure}[]{.3\columnwidth}
		\includegraphics[width=\columnwidth]{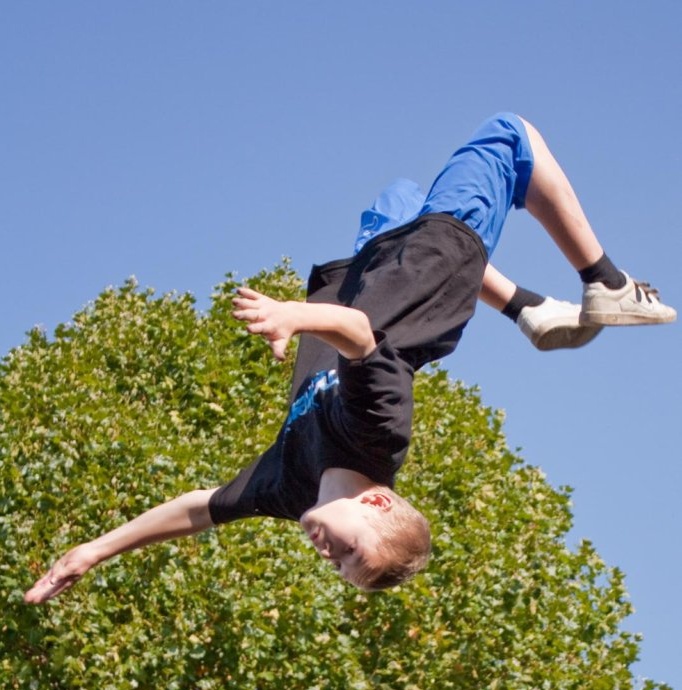}
		\caption*{Image}
	\end{subfigure}
	\begin{subfigure}[]{.3\columnwidth}
		\includegraphics[width=\columnwidth]{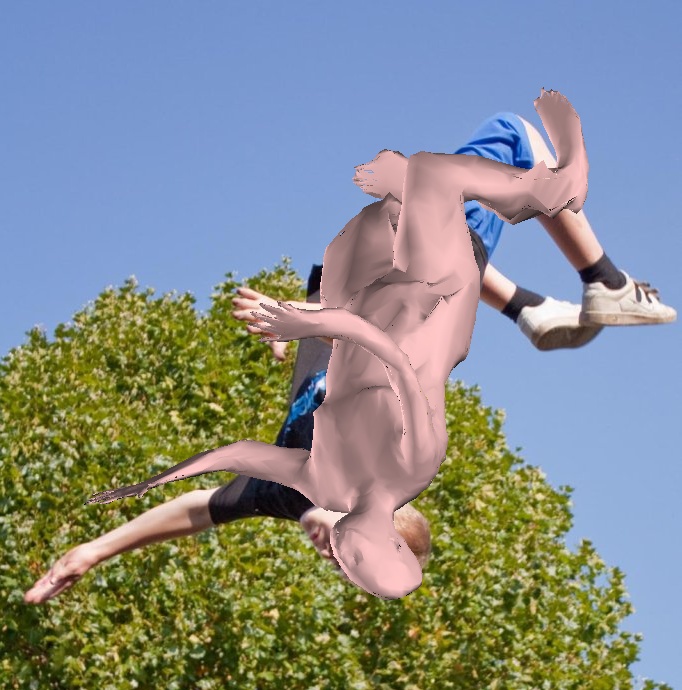}
		\caption*{Non-parametric}
	\end{subfigure}
	\begin{subfigure}[]{.3\columnwidth}
		\includegraphics[width=\columnwidth]{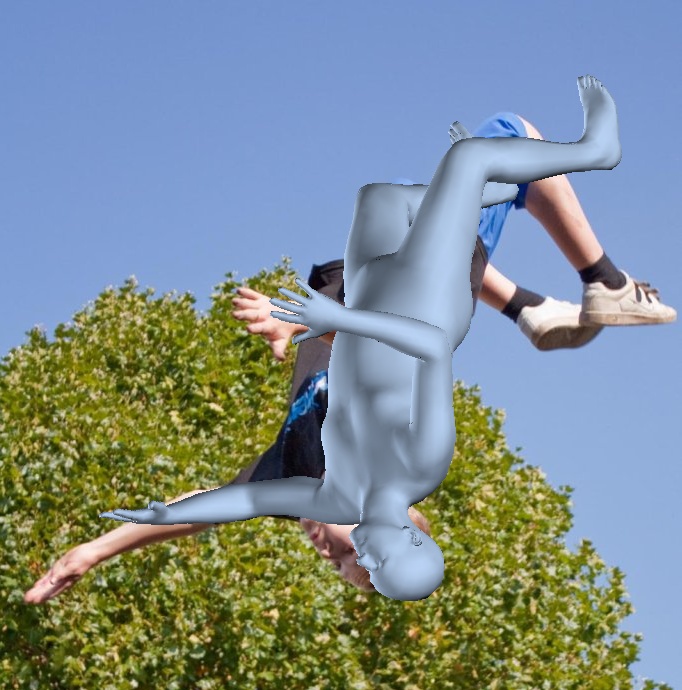}
		\caption*{Parametric}
	\end{subfigure}
	\caption{Examples of erroneous reconstructions. Typical failures can be attributed to challenging poses, severe self-occlusions, or interactions among multiple people.}
	\label{fig:failure}
	\vspace*{-1.5em}	 
\end{figure}

\begin{figure*}[!t]
	\centering
	\begin{subfigure}[]{.15\textwidth}
		\includegraphics[width=\textwidth]{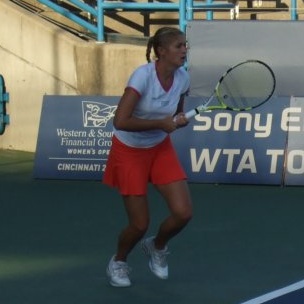}
	\end{subfigure}~
	\begin{subfigure}[]{.15\textwidth}
		\includegraphics[width=\textwidth]{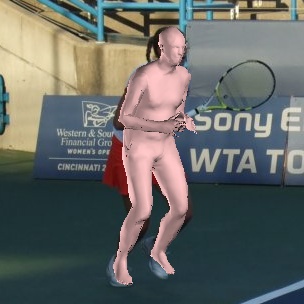}
	\end{subfigure}~
	\begin{subfigure}[]{.15\textwidth}
		\includegraphics[width=\textwidth]{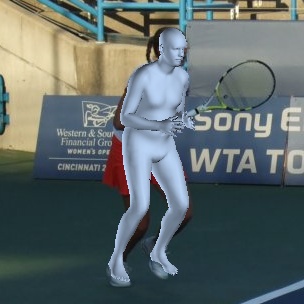}
	\end{subfigure}
	\hfil
	\begin{subfigure}[]{.15\textwidth}
		\includegraphics[width=\textwidth]{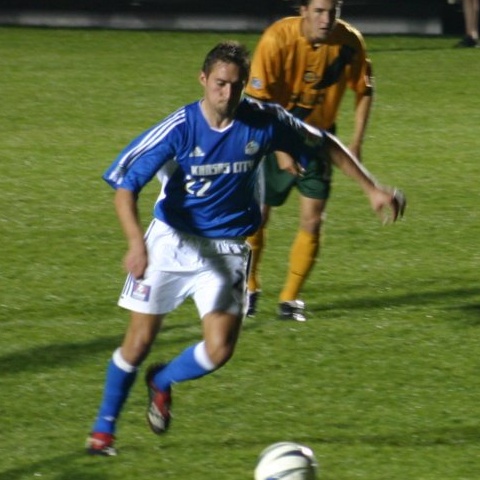}
	\end{subfigure}~
	\begin{subfigure}[]{.15\textwidth}
		\includegraphics[width=\textwidth]{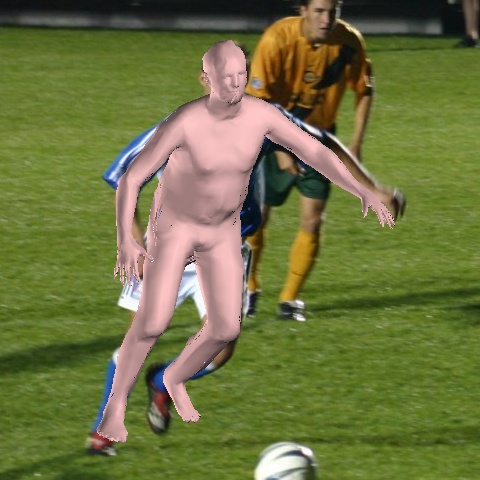}
	\end{subfigure}~
	\begin{subfigure}[]{.15\textwidth}
		\includegraphics[width=\textwidth]{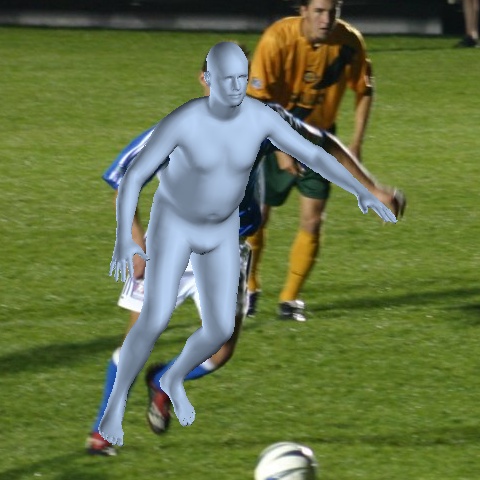}
	\end{subfigure}\\
	\begin{subfigure}[]{.15\textwidth}
		\includegraphics[width=\textwidth]{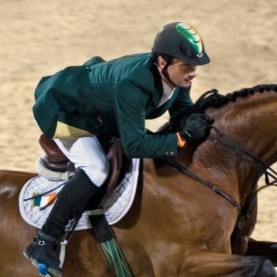}
	\end{subfigure}~
	\begin{subfigure}[]{.15\textwidth}
		\includegraphics[width=\textwidth]{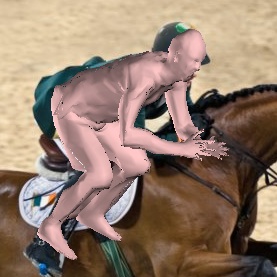}
	\end{subfigure}~
	\begin{subfigure}[]{.15\textwidth}
		\includegraphics[width=\textwidth]{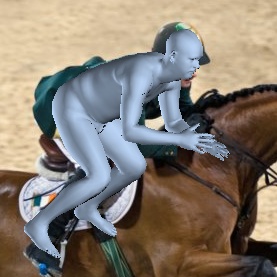}
	\end{subfigure}
	\hfil
	\begin{subfigure}[]{.15\textwidth}
		\includegraphics[width=\textwidth]{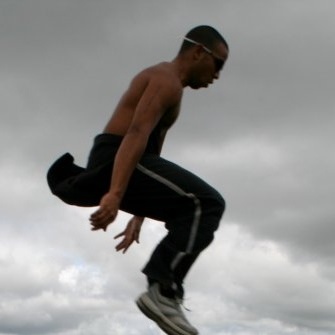}
	\end{subfigure}~
	\begin{subfigure}[]{.15\textwidth}
		\includegraphics[width=\textwidth]{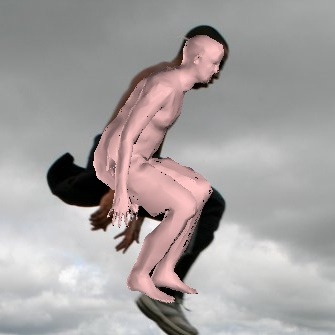}
	\end{subfigure}~
	\begin{subfigure}[]{.15\textwidth}
		\includegraphics[width=\textwidth]{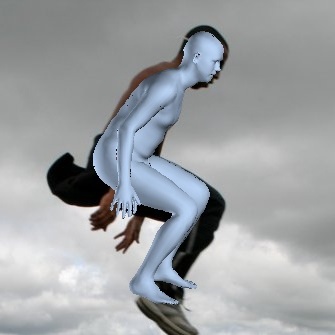}
	\end{subfigure}\\
	\begin{subfigure}[]{.15\textwidth}
		\includegraphics[width=\textwidth]{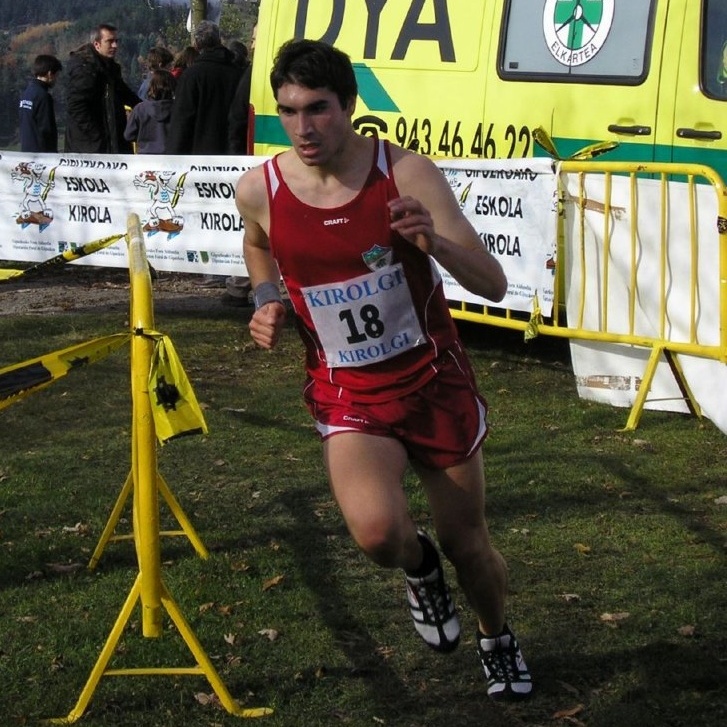}
	\end{subfigure}~
	\begin{subfigure}[]{.15\textwidth}
		\includegraphics[width=\textwidth]{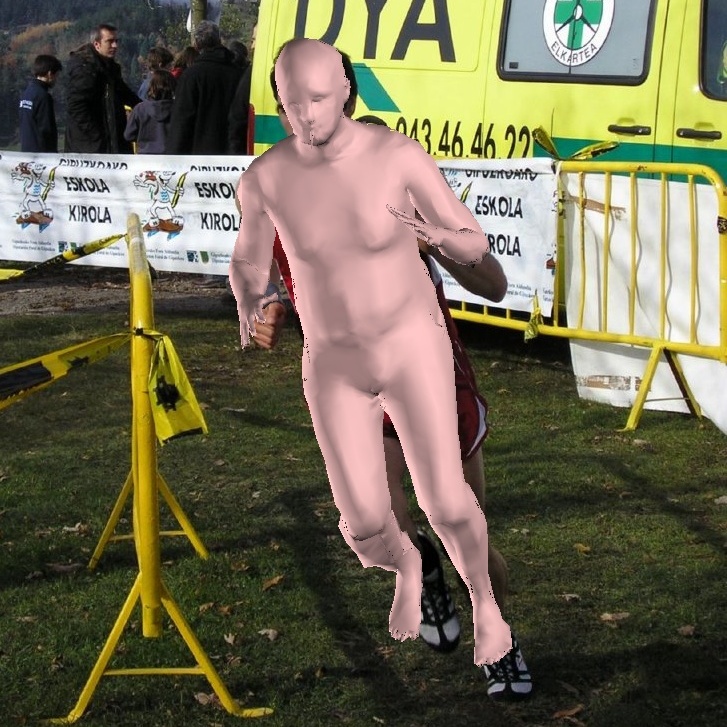}
	\end{subfigure}~
	\begin{subfigure}[]{.15\textwidth}
		\includegraphics[width=\textwidth]{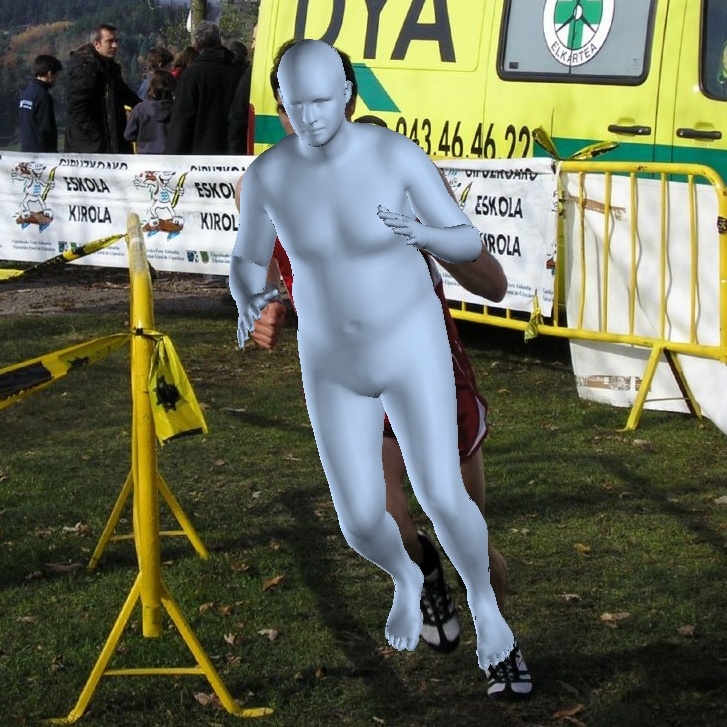}
	\end{subfigure}
	\hfil
	\begin{subfigure}[]{.15\textwidth}
		\includegraphics[width=\textwidth]{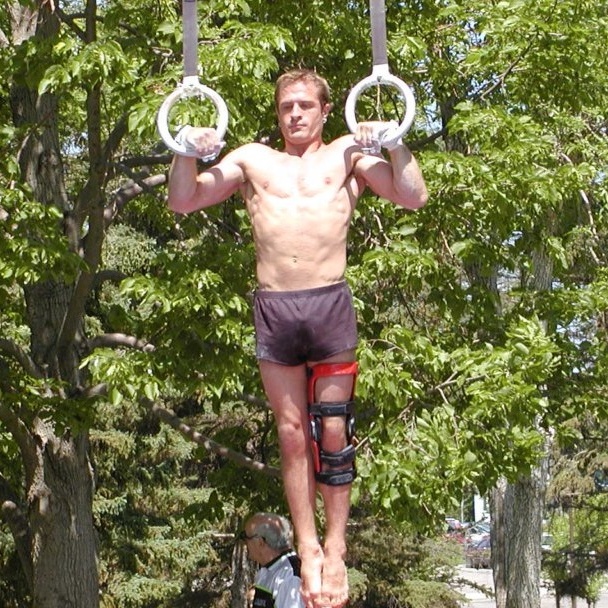}
	\end{subfigure}~
	\begin{subfigure}[]{.15\textwidth}
		\includegraphics[width=\textwidth]{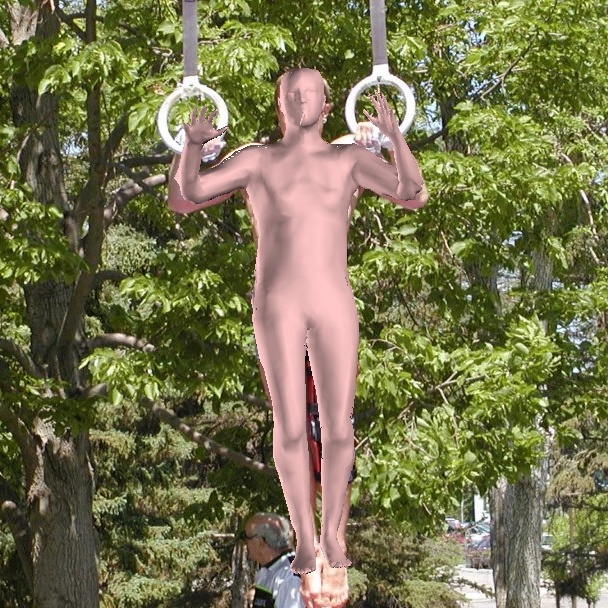}
	\end{subfigure}~
	\begin{subfigure}[]{.15\textwidth}
		\includegraphics[width=\textwidth]{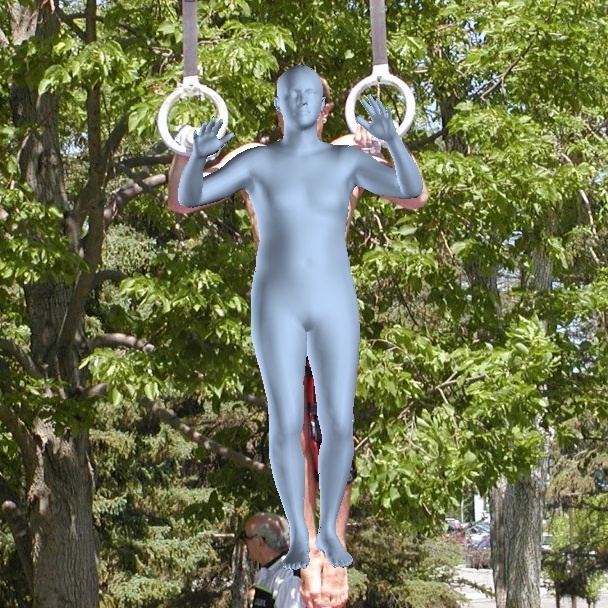}
	\end{subfigure}\\

	\begin{subfigure}[]{.15\textwidth}
		\includegraphics[trim=20mm 20mm 20mm 20mm, clip=true, width=\textwidth]{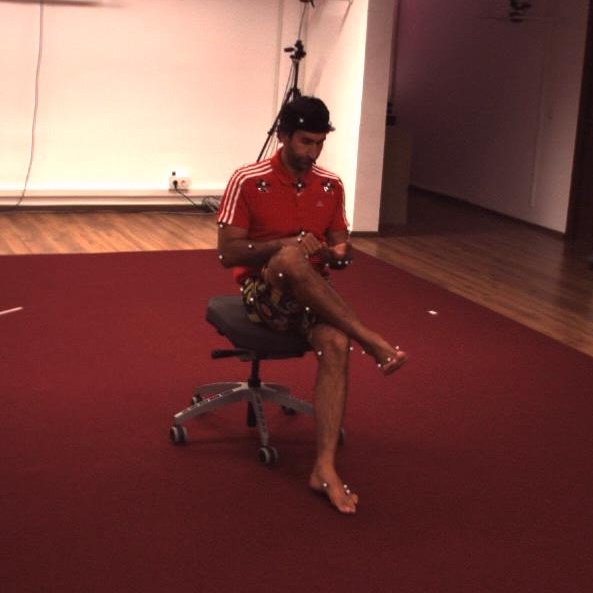}
	\end{subfigure}~
	\begin{subfigure}[]{.15\textwidth}
		\includegraphics[trim=20mm 20mm 20mm 20mm, clip=true, width=\textwidth]{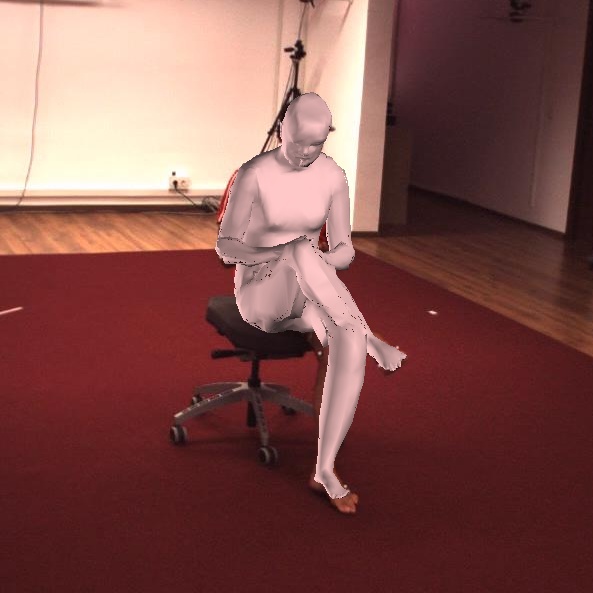}
	\end{subfigure}~
	\begin{subfigure}[]{.15\textwidth}
		\includegraphics[trim=20mm 20mm 20mm 20mm, clip=true, width=\textwidth]{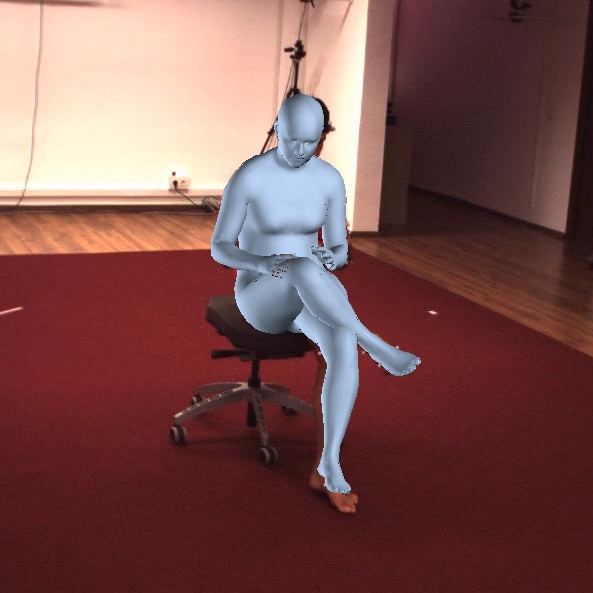}
	\end{subfigure}
	\hfil
	\begin{subfigure}[]{.15\textwidth}
		\includegraphics[trim=20mm 20mm 20mm 20mm, clip=true, width=\textwidth]{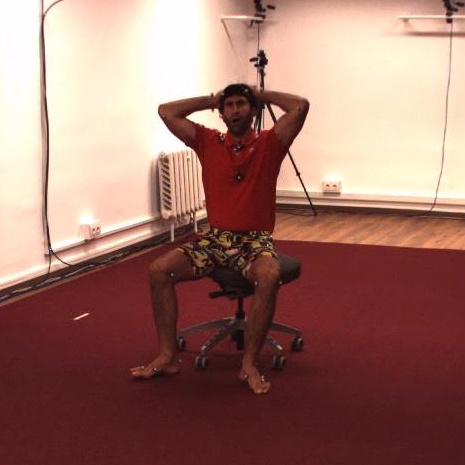}
	\end{subfigure}~
	\begin{subfigure}[]{.15\textwidth}
		\includegraphics[trim=20mm 20mm 20mm 20mm, clip=true, width=\textwidth]{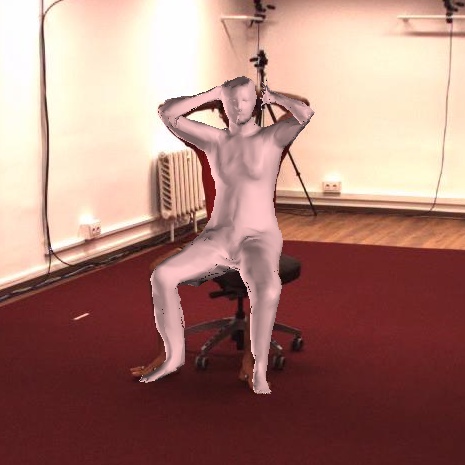}
	\end{subfigure}~
	\begin{subfigure}[]{.15\textwidth}
		\includegraphics[trim=20mm 20mm 20mm 20mm, clip=true, width=\textwidth]{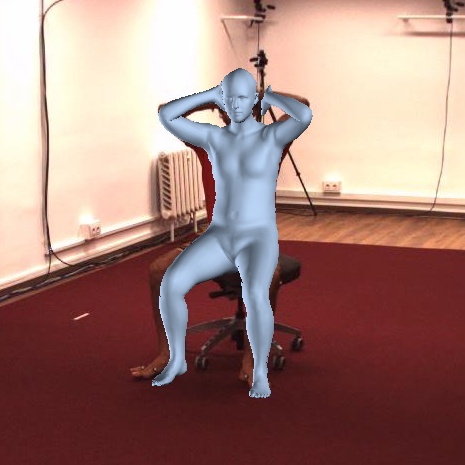}
	\end{subfigure}\\
	\begin{subfigure}[]{.15\textwidth}
		\includegraphics[trim=10mm 10mm 10mm 10mm, clip=true, width=\textwidth]{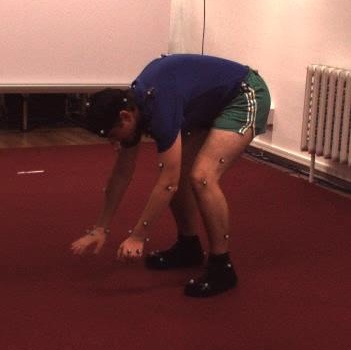}
		\caption*{Image}		
	\end{subfigure}~
	\begin{subfigure}[]{.15\textwidth}
		\includegraphics[trim=10mm 10mm 10mm 10mm, clip=true, width=\textwidth]{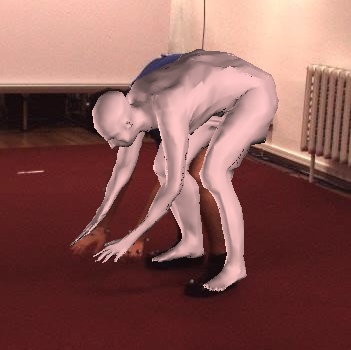}
		\caption*{Non-parametric}		
	\end{subfigure}~
	\begin{subfigure}[]{.15\textwidth}
		\includegraphics[trim=10mm 10mm 10mm 10mm, clip=true, width=\textwidth]{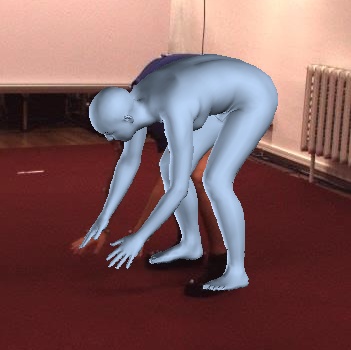}
		\caption*{Parametric}		
	\end{subfigure}
	\hfil
	\begin{subfigure}[]{.15\textwidth}
		\includegraphics[trim=10mm 10mm 10mm 10mm, clip=true, width=\textwidth]{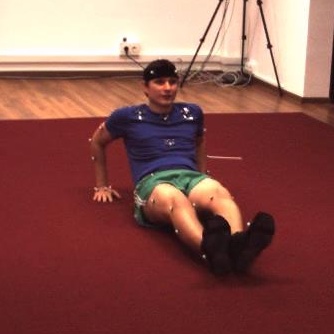}
		\caption*{Image}		
	\end{subfigure}~
	\begin{subfigure}[]{.15\textwidth}
		\includegraphics[trim=10mm 10mm 10mm 10mm, clip=true, width=\textwidth]{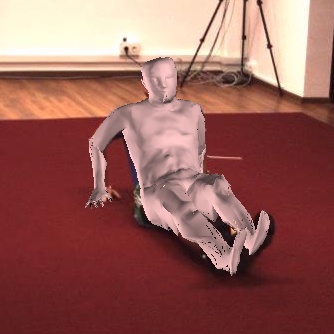}
		\caption*{Non-parametric}		
	\end{subfigure}~
	\begin{subfigure}[]{.15\textwidth}
		\includegraphics[trim=10mm 10mm 10mm 10mm, clip=true, width=\textwidth]{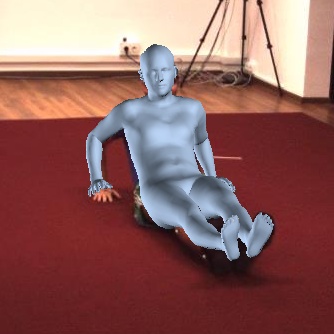}
		\caption*{Parametric}		
	\end{subfigure}\\
\caption{Successful reconstructions of our approach. Rows 1-3: LSP \cite{johnson2010clustered}. Rows 4-5: Human3.6M \cite{ionescu2014}. With light pink color we indicate the regressed non parametric shape and with light blue the SMPL model regressed from the previous shape.}
\label{fig:results}
	\vspace*{-1.0em}
\end{figure*}

\textbf{SMPL from regressed shape}:
Additionally we examine the effect of estimating the SMPL model parameters from our predicted 3D shape. As it can be seen in \tablename~\ref{table:np_smpl}, adding the SMPL prediction, using a simple MLP on top of our non-parametric shape estimate, only has a small effect in the performance (positive in some cases, negative in others). This means that our regressed 3D shape encapsulates all the important information needed for the model reconstruction, making it very simple to recover a parametric representation (if needed), from our non-parametric shape prediction.

\textbf{Comparison with the state-of-the-art}:
Next, we present comparison of our approach with other state-of-the-art methods for 3D human pose and shape estimation. For Human3.6M, detailed results are presented in \tablename~{\ref{table:all}}, where we outperform the other baselines. We clarify here that different methods use different training data (e.g., Pavlakos~\etal~\cite{pavlakos2018learning} do not use any Human3.6M data for training, NBF~\etal~\cite{omran2018neural} uses only data from Human3.6M, while HMR~\cite{kanazawa2018end} makes use of additional images with 2D ground truth only). However, here we collected the best results reported by each approach on this dataset.

Besides 3D pose, we also evaluate 3D shape through silhouette reprojection on the LSP test set. Our approach outperforms the regression-based approach of Kanazawa~\etal~\cite{kanazawa2018end}, and is competitive to optimization-based baselines, e.g.,~\cite{bogo2016keep}, which tend to perform better than regression approaches (like ours) in this task, because they explicitly optimize for the image-model alignment.

\textbf{Qualitative evaluation}:
Figures~\ref{fig:failure} and~\ref{fig:results} present qualitative examples of our approach, including both the non-parametric mesh and the corresponding SMPL mesh regressed using our shape as input. Typical failures can be attributed to challenging poses, severe self-occlusions, as well as interactions among multiple people.

\textbf{Runtime}:
On a 2080 Ti GPU, network inference for a single image lasts 33ms, which is effectively real-time.

\section{Summary}
The goal of this paper was to address the problem of pose and shape estimation by attempting to relax the heavy reliance of previous works on a parametric model, typically SMPL~\cite{loper2015smpl}. While we retain the SMPL mesh topology, instead of directly predicting the model parameters for a given image, our target is to first estimate the locations of the 3D mesh vertices. For this to be achieved effectively, we propose a Graph-CNN architecture, which explicitly encodes the mesh structure and processes image features attached to its vertices. Our convolutional mesh regression outperforms the relevant baselines that regress model parameters directly for a variety of input representations, while ultimately, it achieves state-of-the-art results among model-based pose estimation approaches. Future work can focus on current limitations (e.g., low resolution of output mesh, missing details in the recovered shape), as well as opportunities that this non-parametric representation provides (e.g., capture aspects missing in many human body models, like hand articulation, facial expressions, clothing and hair).

\footnotesize
\noindent
{\bf Acknowledgements:} We gratefully appreciate support through the following grants: 
NSF-IIP-1439681 (I/UCRC), NSF-IIS-1703319, NSF MRI 1626008, ARL RCTA W911NF-10-2-0016, ONR N00014-17-1-2093, ARL DCIST CRA W911NF-17-2-0181, the DARPA-SRC C-BRIC, and by Honda Research Institute.

{\small
\balance
\bibliographystyle{ieee_fullname}
\bibliography{egbib}
}

\end{document}